\documentclass[conference]{IEEEtran}
\usepackage{graphicx}
\usepackage{multirow} 
\usepackage{amsmath}  
\usepackage{amssymb}
\ifCLASSINFOpdf
\else
\fi
\begin{document}
%
\title{M2oE: Multimodal Collaborative \\Expert Peptide Model}

\author{\IEEEauthorblockN{ZengZhu Guo}
\IEEEauthorblockA{school of Information Sciences\\Guangdong University of Finance and Economics\\Guangzhou, China\\
Email: gzz@gdufe.edu.cn}
\and
\IEEEauthorblockN{Zhiqi Ma}
\IEEEauthorblockA{School of Medicine, The Chinese University of Hong Kong, Shenzhen (CUHK-Shenzhen）
Email: $zhiqima@link.cuhk.edu.cn$}
}


%


\title{M2oE: Multimodal Collaborative \\Expert Peptide Model}
\author{\IEEEauthorblockN{Zengzhu Guo$^1$}
\IEEEauthorblockA{School of Information Sciences\\Guangdong University of Finance and Economics\\Guangzhou, China\\
Email: $gzz3383@163.com$}
\and
\IEEEauthorblockN{Zhiqi Ma$^1$}
\IEEEauthorblockA{School of Medicine\\The Chinese University of Hong Kong\\ShenZhen (CUHK-ShenZhen)\\
Email: $zhiqima@link.cuhk.edu.cn$}
}

\maketitle

\begin{abstract}

\textbf{Peptides are biomolecules comprised of amino acids that play an important role in our body. In recent years, peptides have received extensive attention in drug design and synthesis, and peptide prediction tasks help us better search for functional peptides. Typically, we use the primary sequence and structural information of peptides for model encoding. However, recent studies have focused more on single-modal information (structure or sequence) for prediction without multi-modal approaches. We found that single-modal models are not good at handling datasets with less information in that particular modality. Therefore, this paper proposes the M2oE multi-modal collaborative expert peptide model. Based on previous work, by integrating sequence and spatial structural information, employing expert model and Cross-Attention Mechanism, the model's capabilities are balanced and improved. Experimental results indicate that the M2oE model performs excellently in complex task predictions.
Code is available at:https://github.com/goldzzmj/M2oE}

\end{abstract}



%
\IEEEpeerreviewmaketitle

\section{Introduction}
Peptides, which are composed of amino acids, play pivotal roles in the modulation of physiological processes within the body. In contrast to proteins, peptides consist of shorter chains of amino acids\cite{hamley2020introduction}. The prediction of peptide properties entails forecasting their physicochemical characteristics, functions, and biological activities through advanced computational methods that have significantly evolved with the advent of deep learning techniques\cite{serrano2020deepmspeptide}\cite{chen2021sequence}\cite{jiang2023explainable}. Recently, there has been a growing interest in peptides for drug design applications, particularly in the development of antimicrobial and anticancer agents due to the increasing prevalence of antibiotic resistance\cite{wang2021deep}\cite{li2023models}\cite{chen2021xdeep}.

Typically, peptide encoding encompasses both the primary amino acid sequence and its spatial structure. Previous models, including RNN\cite{elman1990finding}, LSTM\cite{hochreiter1997long}, BiLSTM\cite{schuster1997bidirectional}, and Transformer\cite{vaswani2017attention}, indicate that the Transformer architecture is particularly effective in this context. Additionally, peptides can be represented as graph structures, rendering Graph Neural Networks (GNNs) instrumental for capturing molecular spatial information\cite{kipf2016semi}. However, most studies predominantly focus on single-modality data,either sequence or structure,and even contrastive learning techniques often lack a genuine integration of these modalities\cite{liu2023efficient}.

Multimodal models have achieved significant advancements, especially within the AI4Science domain. For instance, GIT-Former\cite{liu2024git} integrates graphical, imaging, and textual information to enhance prediction accuracy in molecular science; meanwhile, Mixture of Experts (MoE) models such as GMoE\cite{wang2024graph} and SwitchTransformer\cite{fedus2022switch} optimize token allocation to improve adaptability. Despite these advancements, multimodal fusion continues to encounter challenges—particularly regarding the refinement of fusion methods for enhanced integration.

To enhance model performance, our M2oE model builds upon previous research by employing a mixed expert framework for embedding, which integrates multiple expert models to achieve more accurate task predictions\cite{jafarpour2013using}\cite{goyal2016multimodal}\cite{dai2023mixture}. This paper presents a multimodal collaborative expert peptide model with the following key contributions:

1. We propose a sequence-structure mixing expert model that addresses the challenge of expert allocation \cite{zhu2024task}.

2.We leverage multimodal characteristics to improve mixed expert representation through interactive attention networks.

3. We utilize learnable weights \(\alpha\) to evaluate the significance of sequence and spatial information across various data distribution scenarios.

\section{METHODS}

\subsection{Benchmark dataset}

The benchmark dataset utilized in this study is derived from Liu et al. \cite{liu2023co}. According to the task division, the datasets encompass classification and regression tasks, which include antimicrobial peptides (AMP) \cite{bhadra2018ampep} and aggregation propensity (AP) \cite{liu2023efficient}. The processing of these two types of datasets aligns with previous work \cite{liu2023co}, having been partitioned into training, validation, and test sets at a ratio of 8:1:1. More detailed information is provided in Table \ref{tab:dataset}.

\begin{table}[h]
\centering
\caption{The classification dataset and the association data set were analyzed, with label 1 in AMPs representing antimicrobial peptides and 0 representing non-antimicrobial peptides.}
\begin{tabular}{cc|c|c}
\hline
\multirow{2}{*}{Dataset}                  & \multirow{2}{*}{Property} & \multicolumn{1}{c|}{Classification} & \multicolumn{1}{c}{Regression} \\ \cline{3-4} 
                                          &                           & \multicolumn{1}{c|}{AMPs}           & \multicolumn{1}{c}{AP}         \\ \hline
\multirow{2}{*}{Train}                    & AMP                       & \multicolumn{1}{c|}{5437}           & \multirow{2}{*}{54159}         \\ 
                                          & non-AMP                   & \multicolumn{1}{c|}{2019}           &                                \\ \hline
\multirow{2}{*}{Validation}               & AMP                       & \multicolumn{1}{c|}{679}            & \multirow{2}{*}{4000}          \\ 
                                          & non-AMP                   & \multicolumn{1}{c|}{252}            &                                \\ \hline
\multirow{2}{*}{Test}                     & AMP                       & \multicolumn{1}{c|}{681}            & \multirow{2}{*}{4000}          \\ 
                                          & non-AMP                   & \multicolumn{1}{c|}{253}            &                                \\ \hline
\multicolumn{2}{c|}{Total}                & \multicolumn{1}{c|}{9321}           & \multicolumn{1}{c}{62159}      \\ \hline
\end{tabular}
\label{tab:dataset}
\end{table}

\subsection{Sequence and Graph encodings}

\begin{figure*}[!t]
\centering
\includegraphics[width=\textwidth]{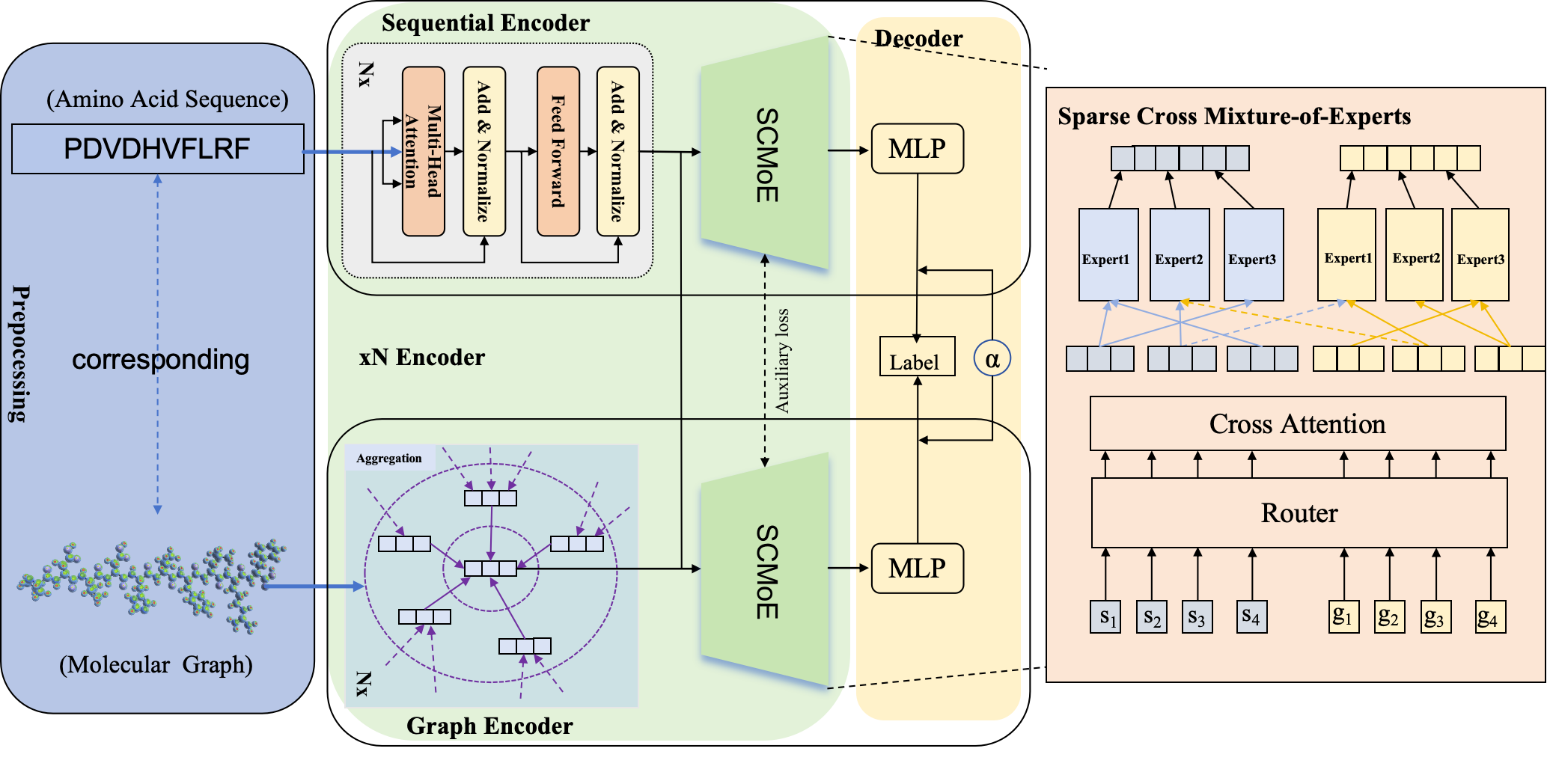}

\caption{The framework of the M2oE. The model is structured with an encoding module and a decoding module, incorporating the interactive attention mechanism in the SCMoE module and MoE token allocation to enhance the comprehensive ability of the M2oE encoding model. Additionally, MoE optimization is achieved through auxiliary loss. The decoding module utilizes MLP and learnable parameters \(\alpha\) from both modes for making predictions.}
\label{fig:framework Co-SCMoE}
\end{figure*}

Peptide sequences $S \subseteq \mathbf{R}^{M}$ and sentence data are similar in that both require word-base embedding and positional identification combination as input. However, the difference is that the division of peptide sequences is based on amino acids and does not require complex tokenizer like natural language. Multi-head Self Attention (MSA) is the core in the Transformer which scores the context and captures various dependencies within the sequence. Feed Forward (FFN) combines with non-linear activate function and additional trainable parameters, further capture non-linear relationships between amino acids and mapped to higher dimension. The sequence encoder output amino acids feature is represented as $s \in S^{M \times d}$, where d is feature hidden dimension.

The peptide molecule is defined as $\mathcal{G}=({\nu}, {\varepsilon })$, where $\nu = \{\nu_{i}\}_{i=1}^{N}$ represents the beads as nodes and $\varepsilon \subseteq  \nu \times \nu $  represents the existence of chemical bonds between the beads as edges. Adjacent matrix $A \in \{0,1\}^{N \times N}$ describes the relationship between nodes and is filled with 0 or 1 based on their corresponding edges, $A_{ij}=1 $, when it is existing connection $(i,j) \in \varepsilon$, otherwise $A_{ij}=0 $. Given feature adjacent X and join with adjacent A, GCN\cite{kipf2016semi} leverages relative edges and nodes attribute to learn latent representation of the node. One layer graph convolutional encoder represents as follow:
\begin{equation}
    X^{(l+1)}=f_{GCN}(A,X^{(l)};W^{(l)})=
    \sigma (\hat{A}X^{(l)}W^{(l)}),
\end{equation}

where $f_{GCN}$ is GCN encoder function, $\widetilde{A} = A + I$ add diagonal matrix to keep and transmit the information of the node itself, $\hat{A}= {D}^{-\frac{1}{2}}\widetilde{A}{D}^{-\frac{1}{2}}$ is to normalize the adjacency matrix. $W^{(l)}$ represent the learnable weight matrix of the l-layer of the model and $\sigma$ is a non-linear activate function LeakyRelu. The Initial values $X^{(0)}$ are randomly initialized using a normal distribution and the final output by GCN is represented as $ X \in \mathbb{R}^{N\times D}$ where D donates each node embedding dimension.
\subsection{Sparse Cross Mixture of Experts}

As shown in Figure 1, the parallel Transformer and SAGEGraph capture the primary peptide sequence information and the secondary molecular structure information. However, sequence information and structural information can represent and complement each other. Therefore, we have designed a sparse interaction mixed expert system (SCMoE) fusion module.

The SCMoE model contains C sequence mixing experts and graph mixing experts, which can learn from tokens routed by different types of data through the expert network. In particular, the interactive attention network possesses the ability to focus on different modalities directly, endowing the mixing experts with stronger representational capabilities through this multimodal alignment approach. Specifically, the routing network is controlled by a learnable matrix \(W^{r} \in \mathbb{R}^{d \times C}\), which calculates the similarity between each token and the mixing experts, and assigns them to the \(topk\) most similar experts. The formula \ref{Gate function} shows this allocation method, where \(X_{ij}\) represents assigning the i-th token to the j-th expert with a coefficient of \(\alpha\). 

However, using the Top k allocation method alone may result in some tokens never being assigned to experts, thus reducing the expressive power of the expert system\cite{shazeer2017outrageously}. To address this issue, a stochastic variable sampled from the standard normal distribution is added, allowing tokens ranked after K to also have a chance for allocation.
\begin{equation}
\begin{aligned}
\
Router(X_{i}) &= Topk( \alpha_{j} X_{ij} + \textit{N}(0,1) \cdot Softplus(X_{ij}W_{noise}))  ; \\
  \alpha_{j} &=  \frac{X_{ij}W^{r}}{ {\textstyle \sum_{j=0}^{topk}}X_{ij}W^{r}} 
    \label{Gate function}
\end{aligned}
\end{equation}

Among them, $W_{noise}\in\mathbf{R}^{d \times C}$ are learnable parameters and \(Softplus(\cdot)\) is a nonlinear activation function can prevent the problem of vanishing gradients.

The peptide sequence is composed of multiple amino acid symbols, so each character can be used as a local feature. The combination of local features assigned to the mixed experts implicitly expresses certain characteristics of the peptide sequence. However, relying solely on single-modal information makes it difficult to directly learn the implicit characteristics of peptides. Therefore, the Cross-Attention (CRA) is proposed to improve the MoE\cite{liu2024git}. It can align similar characteristics between modalities while also drawing away different characteristics. Specifically, it can be represented as follows:

where \( F_{seq}, F_{gra} \) denote features from the sequence encoder and graph encoder, and \( d_k \) is the scaling factor respectively. Subsequently, we exchange the queries \( Q \) of the two modalities for spatial interaction:
\begin{equation}
\begin{aligned}  
    F_{fgra}&= \text{Softmax}\left(\frac{Q_{seq} K_{gra}^\top}{d_k}\right)V_{gra}, \\
    F_{fseq}&= \text{Softmax}\left(\frac{Q_{gra} K_{seq}^\top}{d_k}\right)V_{seq}, \\
    \label{Cross Attention function}
\end{aligned} 
\end{equation}

Subsequently, the cross-attention matrices of the two modalities are transformed and updated. The new sequence features are composed of graph node features and their corresponding attention coefficients. The updated interactive features also need to be allocated to different experts, similar to the formula\ref{Gate function}. 
Therefore, the updated sequence features can be integrated into the routing network through the operation \( F_{seq}^{new} = \text{Concat}(F_{seq}, F_{fseq}) \), as do the graph node features.

\subsection{Fusion Module And Loss}

The antimicrobial peptide prediction task is conducted based on the sequence and its spatial structure. Our designed SCMoE module ensures the expression of characteristics of each modality and enhances the expression of potential features with the help of information from another modality. Therefore, the final fusion module only needs to utilize the nonlinear capability of MLP to capture the correlation between features and map them to the classification space of antimicrobial peptides. 
Traditional methods often involve combining multiple output results using fixed weights, but this approach is limited in that it is difficult to assess the importance of sequence and spatial information for prediction under different data distribution scenarios. As shown in formula \ref{Fusion function}, we employ learnable weights \(\alpha\) to measure this importance.
\begin{equation}
\begin{aligned}  
    \hat{y} = \sigma(\alpha MLP_{1}(Z_{seq}) + (1-\alpha)MLP_{2}(Z_{gra}))
    \label{Fusion function}
\end{aligned} 
\end{equation}

Among them, \(\sigma\) is Sigmoid function, mapping predictive data into the probability space. \(Z_{seq} ,Z_{gra}\)are embeddings of the output from the sequence encoder and the graph encoder.

The routing network assigns tokens to experts based on the gating method, but this approach can sometimes lead to load imbalance issues, where one expert receives the majority of tokens, thereby degenerating into a single-expert model. Therefore, a strategy designed to ensure that each expert has an equal probability of being selected is formulated as shown in Equation \ref{Loss load function}. 
On the other hand, the capabilities of each expert are different, and the routing network tends to allocate tokens to the few experts with stronger capabilities, leaving the remaining experts idle, which similarly leads to load imbalance issues. As shown in Equation \ref{Loss importance function}, the \(CV(\cdot)\) function measures the degree of discreteness of expert importance, combined with fixed hyperparameter \(\omega_{imp}\) to control the similar abilities of different experts.
\begin{equation}
\begin{aligned}  
    L_{load} = \sum_{i=1}^{C} \left( \frac{n_i}{\sum_{j=1}^{C} n_j} - \frac{1}{C} \right)^2
    \label{Loss load function}
\end{aligned} 
\end{equation}

\begin{equation}
\begin{gathered}
    L_{importance} = \omega_{imp} \cdot CV( {\textstyle \sum_{x\in X}Router(x)} ) \\
    CV(X) = \frac{\sigma_{x}}{\mu_{x}}
    \label{Loss importance function}
\end{gathered}   
\end{equation}

Among them \(\sigma_{x}\)and \(\mu_{x}\) are the variance and mean of data X.

Finally, the error between the predicted values and the true labels is calculated using the Binary Cross Entropy (BCE), and this is added to the balanced loss function of Mode of Expertise (MoE) regarding load and importance as the total optimization objective.

\begin{equation}
\begin{aligned}  
    L = BCE(y,\hat{y}) + L_{Load} + L_{importance}
    \label{final Loss function}
\end{aligned} 
\end{equation}

\section{RESULTS AND DISCUSSION}

We propose the M2oE model, which effectively balances and integrates sequence and structural features for downstream tasks. This model encompasses three types: sequence, graph, and hybrid models. The sequence and graph models are single-modality frameworks evaluated on classification (AP) and regression (AMP) datasets. For the sequence model, we employ Transformer and SwitchTransformer architectures, while the graph model utilizes GCN, GAT, GraphSAGE, and GMoE \cite{fedus2022switch, wang2024graph}. The hybrid model incorporates Repcon, weighted fusion M2oE methods, concatenation techniques, as well as our final approach.

Table \ref{tab:Compare result} illustrates that the sequence model demonstrates superior performance on the AP dataset; notably, SwitchTransformer achieves an impressive \(R^2\) of 95.1\%. Conversely, on the AMP dataset, GraphSAGE leads with an accuracy of 84.7\%. These findings suggest that single-modality models excel when a dataset is biased towards one modality but encounter challenges when it favors another.

The M2oE model synergistically combines the strengths of both sequence and graph models to achieve remarkable performance across both datasets: \(R^2 = 0.951\) on AP with minimal MAE and MSE values (3.68E-2 and 2.21E-3), alongside an accuracy of 86.2\% on AMP—surpassing baseline results.

While MoE enhances performance in single modalities independently without benefiting other modalities directly; therefore we implement Cross-Attention to ensure balanced improvements across modalities. Ablation experiments presented in Table \ref{tab:Ablation experiment result} corroborate this assertion. Ultimately demonstrating that M2oE achieves optimal results by improving upon baseline metrics by 0.9\%.

\begin{table}[h]
\centering
\caption{Ablation experiment result on the AP datasets.}
\begin{tabular}{cccc}
\hline
Variants                 & MAE              & MSE              & $R^{2}$             \\ \hline
M2oE without CRA nor MoE & 3.96E-2          & 2.57E-3          & 0.942          \\
M2oE without CRA         & 3.74E-2          & 2.27E-3          & 0.949          \\
M2oE without MoE         & 3.79E-2          & 2.38E-3          & 0.947          \\
M2oE                     & \textbf{3.68E-2} & \textbf{2.21E-3} & \textbf{0.951} \\ \hline
\end{tabular}
\label{tab:Ablation experiment result}
\end{table}

\begin{table}[h]
\caption{Comparison with state-of-the-art methods, including sequence models, graph models, and mixed models.}
\begin{tabular}{c|c|ccc|cc}
\hline
\multirow{2}{*}{Type}     & \multirow{2}{*}{Model} & \multicolumn{3}{c|}{AP}   & \multicolumn{2}{c}{AMP} \\ \cline{3-7} 
                          &                        & MAE     & MSE     & $R^{2}$    & ACC   \\ \hline
 \multirow{2}{*}{Sequence} 
                          & Transformer            & 3.81E-2 & 2.33E-3 & 0.947 & \underline{0.813} \\
                          & SwitchTransformer\cite{fedus2022switch}      & \underline{3.65E-2} & \underline{2.15E-3} & \underline{0.951} & 0.808 \\ \hline
\multirow{4}{*}{Graph}    & GCN                    & 4.27E-2 & 3.02E-3 & 0.932 & 0.834 \\
                          & GAT                    & 4.40E-2 & 3.22E-3 & 0.928 & 0.843 \\
                          & GraphSAGE              & 3.84E-2 & 2.36E-3 & 0.947 & \underline{0.847} \\
                          & GMoE  \cite{wang2024graph}                 & \underline{3.82E-2} & \underline{2.35E-3} & \underline{0.947} & 0.837 \\ \hline
                          
\multirow{3}{*}{Mixture}  & Repcon(Avg)            & 3.83E-2 & 2.24E-3 & 0.947 & 0.831 \\
                          & M2oE(WS)               & 3.74E-2 & 2.29E-3 & 0.949 & 0.820 \\
                          & M2oE(Concat)           & 3.73E-2 & 2.26E-3 & 0.949 & 0.824 \\
                          & M2oE(Parallel)         & \textbf{3.68E-2} & \textbf{2.21E-3} & \textbf{0.951} & \textbf{0.862} \\ \hline
\end{tabular}
\label{tab:Compare result}
\end{table}

\section{Conclusion}

In this paper, we propose a multimodal collaborative expert peptide model, which integrates sequence and spatial structural information, utilizes a sparse mixed expert model, and takes into account the characteristics under different data distributions. Experimental results show that the M2oE model performs well in complex task prediction, and uses multimodal methods to solve problems that may arise in unimodal scenarios. Finally, we use ablation experiments to demonstrate the effectiveness of each module. In future work, we can also consider connecting the multimodal expert model to more complex tasks, such as peptide generation tasks, etc.

 \bibliographystyle{IEEEtran}
\bibliography{reference}

\end{document}